\documentclass{zoooo}
\usepackage{tabularx}

\title{Toward Social World Models: \\Emergent Stories and Playable Cinematic Worlds}

\author{Zoooo team}
\affiliation{Banana Lab, HKUST, Stanford University}
\abstract{
We envision \textbf{social world models} that sustain persistent,
evolving virtual societies, where autonomous characters
accumulate experiences, develop relationships, and create
stories that people can watch, enter, and influence.
Realizing this vision requires preserving the social causes
and consequences that connect events, relationships, and
user actions over time.
We present \textbf{Zoooo-WM}, a step toward this vision built
around three key capabilities: long-horizon social evolution,
explicit state control, and multi-agent interaction.
These capabilities support continuously unfolding social
histories from which meaningful stories can be discovered,
experienced from different perspectives, and shaped through
user participation.
By bringing emergent storytelling, cinematic visual quality,
and playability together, Zoooo-WM supports character-driven
story development and visualization for filmmaking, evolving
social encounters and controllable experiences for games,
and interactive cinema in which audiences can enter a story
and influence its course.
\par\vspace{3mm}
{\small
\noindent\textsf{\textbf{Official page:}}\enspace
\url{https://research.zoooo.ai/Zoooo-WM}\par
}
}

\begin{document}
\maketitle

\par\addvspace{0.5em}
\noindent
\begin{minipage}{\textwidth}
    \centering
    \includegraphics[width=0.9\linewidth]
        {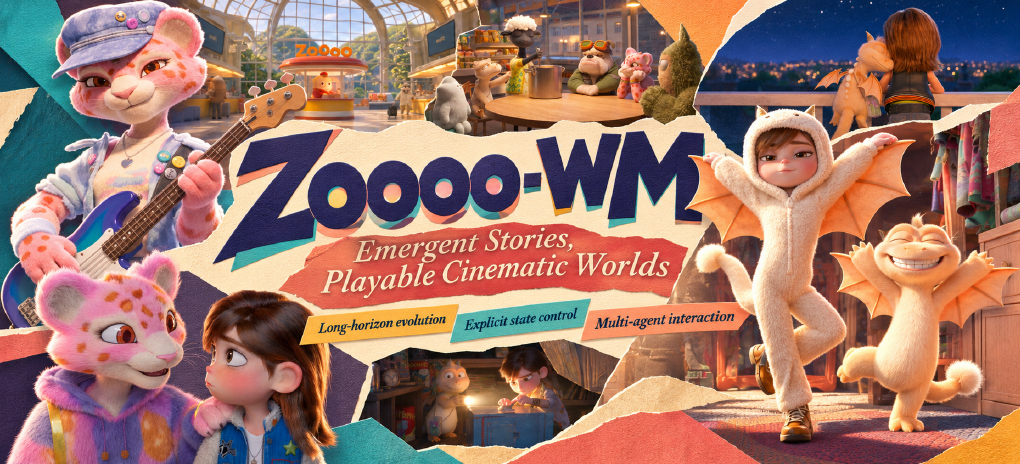}

    \captionsetup{
        font=small,
        skip=6pt,
        justification=justified,
        singlelinecheck=false
    }
    \captionof{figure}{
        \textbf{Zoooo-WM: Emergent Stories, Playable Cinematic Worlds.}
        Zoooo-WM advances toward social world models through
        long-horizon evolution, explicit state control, and
        multi-agent interaction.
        These capabilities enable cinematic storytelling and
        playable experiences for film, games, and interactive cinema.
    }
    \label{fig:teaser}
\end{minipage}
\par\addvspace{0.75em}

\section{Introduction}
Generative AI opens a path toward virtual worlds that people can watch, enter, and influence~\cite{videoworldsimulators2024,hadsell2026genie4,worldlabs2026atlas,park2023generative,pan2_2026,sun2025worldplay,team2026advancing,xiang2025pan,bruce2024genie,openai2025sora2}. 
From a brief premise and a set of character descriptions, we envision virtual societies in which characters pursue goals, accumulate experiences, and create stories through their interactions. These stories can be discovered from different perspectives, realized with cinematic visual quality, and shaped by users who participate in their unfolding. We explore this vision through \textbf{social world models}: persistent, evolving worlds that connect autonomous social behavior with storytelling and play.

Our motivation is to make generated worlds sustain the social causes and consequences that give stories meaning. An encounter can change a relationship; a decision can affect what another character knows; a user intervention can alter the course of later events~\cite{mccoy2011comme,brenner2010creating,mateas2005structuring}. Supporting such experiences requires continuity across time, interaction among characters with distinct personality settings, and a world state that can be deliberately changed. A simulated society also produces many everyday events, creating a need to discover meaningful narrative arcs and choose how to present them~\cite{ryan2018curating,kreminski2019felt}. These requirements motivate an approach organized around long-horizon social evolution, explicit state control, and multi-agent interaction.

Prior research provides complementary foundations. Interactive generative video world models advance user control and sustained interaction with generated environments~\cite{hadsell2026genie4,team2026advancing,zhang2025matrix,huang2026solarwm,zhang2026echowm,gao2026infinite}. Coding World Models further connect executable state with generative rendering, making world behavior available for explicit inspection and programmatic control~\cite{zhang2026matrix,chen2026code,huang2026programmable,wang2026cwm2}. These directions provide building blocks for the visual, interactive, and executable aspects of a social world. Generative Agents offers a closely related foundation: its Smallville environment demonstrates how memory, reflection, and planning support evolving relationships and coordinated social behavior, with users able to observe and intervene~\cite{park2023generative,vezhnevets2023generative,park2024generative,kolluri2025finetuning}. 
Multi-agent interaction is central to this foundation\cite{yang2411oasis,piao2025agentsociety,ashery2025emergent,zhou2024sotopia}:
agents communicate, respond to one another, and adapt their
plans as social encounters unfold.
These interactions enable information diffusion, relationship
formation, and collective coordination, allowing shared events
to emerge from individual decisions~\cite{park2023generative}.
Our emphasis is on connecting such evolving societies to narrative discovery and cinematic participation.
The central question is how a shared social history can become a source of stories that people can follow from different perspectives, experience visually, and influence through their actions.

In this report, we present \textbf{Zoooo-WM}, a social world model that connects persistent social simulation, narrative discovery, and explicit world control to create emergent stories and playable cinematic experiences. The system begins with a story premise, character profiles, and an initial world configuration. Agents pursue goals, draw on prior experiences, and interact with shared objects and one another, accumulating a history that informs subsequent decisions and relationships. Following the perspective of discovering narratives within simulated activity~\cite{kreminski2023authoring,leong2022automated}, a Story Agent identifies connected events, conflicts, and delayed consequences, then selects viewpoints and organizes the material into scripts and storyboards. The same social history can yield multiple stories, depending on which characters and developments are followed. This connects autonomous social evolution to deliberate choices about what to show and how to tell it. Coding Agent Orchestration translates world state, selected events, and user requests into executable scenes with explicit character, object, action, and camera states. Users can follow a selected story or intervene from a saved checkpoint through direct controls or language-based requests. We frame playability as the ability to enter a scene, act within it, and change events with consequences that persist. The interaction loop is designed to carry execution outcomes back into the social simulation, informing later agent decisions and event histories. Structure videos from this execution layer, together with appearance-conditioned reference frames, guide a pretrained video renderer to produce cinematic sequences. Our rendering work focuses on integrating pretrained generative models and accelerating inference for streaming output.

This combination of playability and cinematic visual quality opens applications across film, games, and interactive cinematic experiences. For film and television, the pipeline supports exploring character-driven story possibilities, discovering narrative arcs, and visualizing selected scenes for story development and previsualization. For games, autonomous characters and executable world state provide a foundation for player control, social encounters, and situations that develop through interaction. Interactive cinematic experiences bring these possibilities together: audiences can watch a story unfold, enter a character's perspective, or intervene at a selected moment and explore an alternative continuation. The same underlying society can thus support both cinematic storytelling and participatory play.

This report presents the system's organizing principles and examines its
capabilities and limitations through qualitative demonstrations.
We organize our contributions around three themes:
\begin{enumerate}
    \item \textbf{Long-horizon social simulation and narrative discovery.}
We support long-horizon social evolution through an autonomous agent
community that continually generates new events and storylines through
interaction. A Story Agent selects and distills compelling developments
from this ongoing simulation into coherent narratives, choosing viewpoints
for their presentation while the underlying community continues to evolve
beyond any individual story.

    \item \textbf{Explicit, executable world control through coding-agent orchestration.}
    We realize explicit state control by translating social state,
    narrative intent, and user interventions into executable structure.
    This grounds user participation in concrete world changes and supports
    state persistence, direct control, and story branching.

    \item \textbf{Integrated cinematic and playable experiences built with pretrained generative models.}
    We connect multi-agent interaction with executable control and
    pretrained cinematic rendering, turning exchanges among characters
    with distinct goals and perspectives into stories that users can
    watch and influence. This integration supports cinematic and playable
    experiences for filmmaking, games, and interactive cinema.
\end{enumerate}

\section{System Overview}
\label{sec:system_overview}
\begin{figure*}[!t]
    \centering
    \includegraphics[width=\textwidth]
        {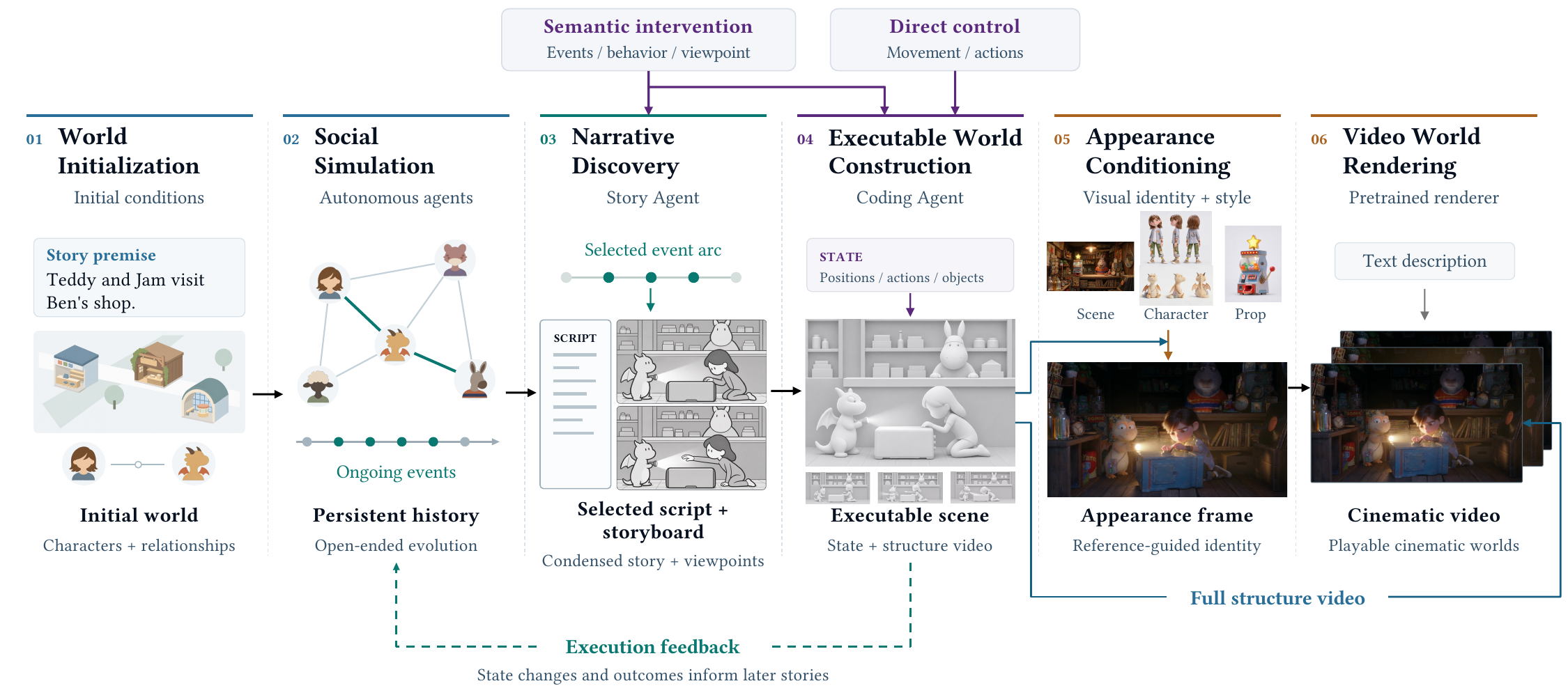}
    \caption{
        \textbf{Overview of Zoooo-WM.}
        The system connects persistent social simulation, narrative discovery,
and executable world control with pretrained cinematic rendering.
Users participate through direct control and semantic intervention.
The feedback loop is designed to carry execution outcomes back into
social simulation, shaping subsequent events and stories.
    }
    \label{fig:system_overview}
\end{figure*}

Figure~\ref{fig:system_overview} presents the overall architecture
of Zoooo-WM, which connects an evolving social world to cinematic
and playable experiences through six stages.
Together, these stages transform initial world and character
settings into persistent social histories, selected narratives,
executable scenes, and generated video.
Users participate through direct control and semantic intervention,
while executed state changes and interaction outcomes feed back
into the social simulation to shape subsequent events and stories.

\subsection{World Initialization}
\label{sec:overview_initialization}
A story premise, character profiles, initial relationships, and a world configuration establish the starting conditions of the simulation. These inputs define who inhabits the world, what motivates them, and which places and objects they can interact with. Scene, character, and key-prop reference images specify the desired appearance for later visual generation.

\subsection{Social Simulation}
\label{sec:overview_social_simulation}
Agents act according to their goals, memories, and observations, interacting with shared objects and one another. Persistent world state and event histories carry the consequences of these actions into subsequent decisions and relationships. This supports long-horizon social evolution and produces the events from which stories can be discovered.

\subsection{Narrative Discovery}
\label{sec:overview_narrative_discovery}

Story Agents identify meaningful event arcs within the accumulated social simulation history, condense the selected events, and choose characters and viewpoints to follow. It organizes this material into a script and storyboard, specifying what to present and from which perspective. The same history can therefore support multiple stories, each highlighting different developments and experiences.

\subsection{Executable World Construction}
\label{sec:overview_world_construction}
Coding Agents translates the script, storyboard and its associated world state into executable scenes. It orchestrates character actions, object interactions, and camera trajectories within a deterministic execution environment. This stage maintains explicit scene state and produces a structure video capturing the intended composition, motion, and timing. The underlying executable scene also supports direct user control.

\subsection{Appearance Conditioning}
\label{sec:overview_appearance_conditioning}
We combine scene, character, and prop references with the first structure frame to establish an appearance-conditioned initial frame. The references specify visual identity and style, while the structure frame guides composition, poses, and object placement. This establishes the visual starting point for rendering the selected story.

\subsection{Video World Rendering}
\label{sec:overview_video_rendering}
A pretrained video model receives the appearance-conditioned initial frame, the corresponding structure video, and a text description to generate a cinematic sequence. Structure conditioning connects visual generation to the executed scene, while inference-time acceleration supports streaming output. This stage gives the simulated and orchestrated story its cinematic visual presentation.

\paragraph{User Interaction and Social Feedback.}
Users can watch selected stories or participate through direct control, such as character movement, camera control, and immediate actions, and semantic intervention, such as changing a character's behavior, modifying an event, or choosing another viewpoint. The Coding Agent translates world-changing requests into executable actions or state changes; the Story Agent handles requests about narrative selection and perspective.

The interaction loop is designed to feed executed changes and interaction outcomes back into the social simulation. As these effects become available to agents through their observations and experiences, they inform later decisions, relationships, and event histories. The Story Agent can then select stories from the resulting continuation. Changes to viewpoints or appearance offer another presentation of the world; interventions that affect its state provide a basis for new story trajectories.
\section{Social Simulation and Narrative Discovery}

Zoooo-WM sustains an evolving virtual society through four coordinated modules: a world state and simulated environment module, a character perception and autonomous decision-making module, an embodied interaction module, and a persistent history and narrative discovery module. By grounding open-ended language-model reasoning in explicit spatial state and perceptual boundaries, the system drives long-horizon autonomous social evolution and multi-perspective story emergence while providing a structured foundation for downstream executable scene construction and external state control.

\begin{figure}[htbp]
    \centering
    \includegraphics[width=\linewidth]{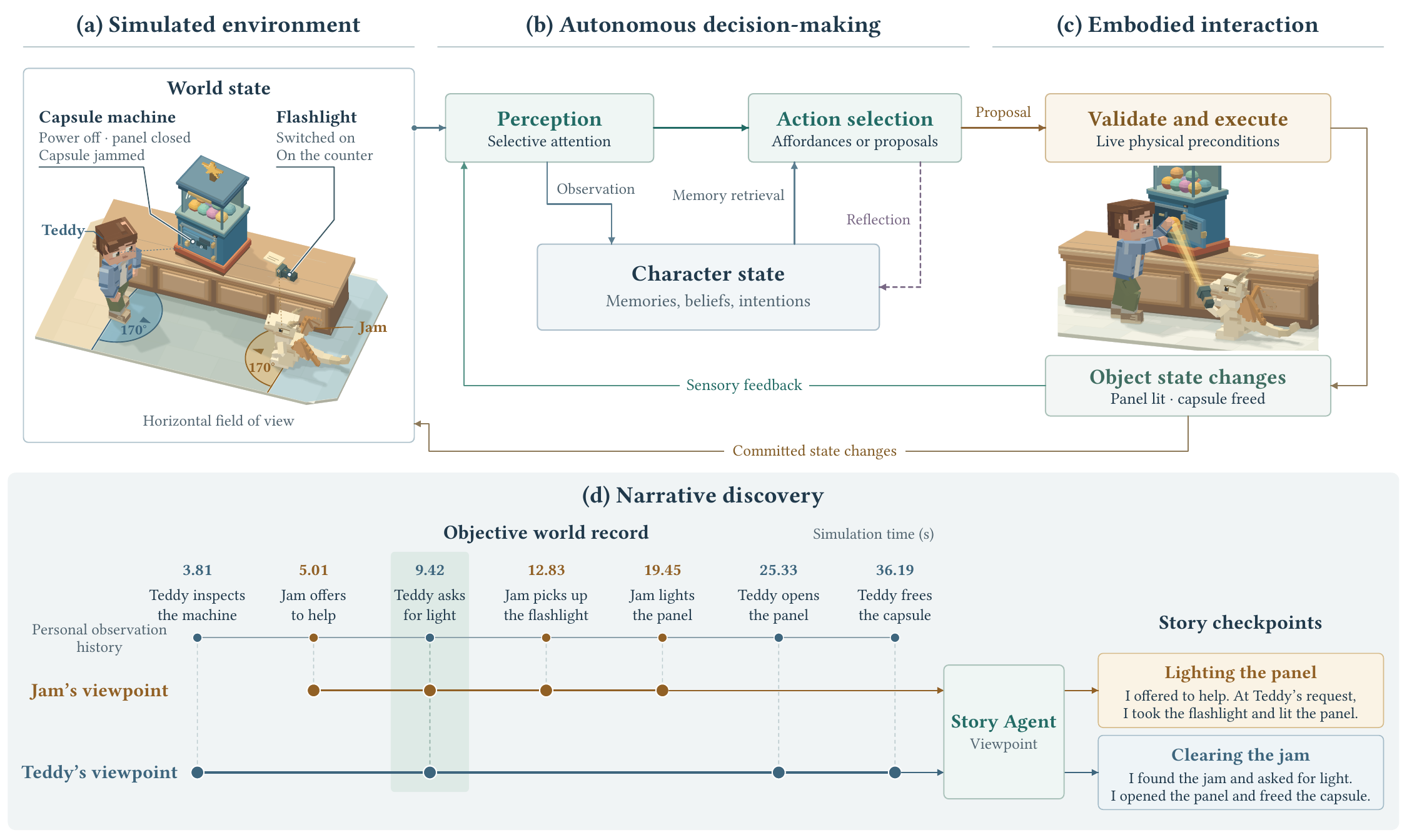}
    \caption{
    \textbf{Social simulation and narrative discovery.}
    (a) The environment maintains explicit object states, which characters perceive within their respective fields of view. (b) Observations update each character's memories, beliefs, and intentions, informing action selection through memory retrieval and reflection. (c) Action proposals are validated against live physical preconditions and executed, returning sensory feedback and committing state changes to the world. (d) From the objective world record and each character's personal observation history, the Story Agent extracts viewpoint-specific story checkpoints.
    }
    \label{fig:swm-social-simulation}
\end{figure}

\subsection{World State and Simulated Environment}

The simulated environment represents characters, props, and environmental fixtures as stateful objects within a shared three-dimensional space. Each object maintains semantic attributes, geometric bounds, and spatial topological relationships, forming a structured environment for social activity~\cite{weyns2007environment}. The entire world progresses under a unified, continuous simulation clock, with the spatial positions, topological attachments, and semantic states of all objects updating continuously over the course of the simulation. Once triggered, continuous environmental processes evolve independently over time according to their underlying spatial relationships and physical conditions, feeding resulting state transitions and sensory signals back into the environment in real time to provide dynamically evolving external conditions for character perception and decision-making.

\subsection{Perception, Memory, and Autonomous Decision-Making}

Realistic interpersonal dynamics build upon information differences across individuals~\cite{ryan2015toward}. The system restricts each character's decision context to information acquired through personal experience. Characters perceive the world through multimodal sensory channels: vision is constrained by field of view and physical occlusion, hearing is governed by propagation distance and distinguishes conversational roles, close-range inspection relies on physical proximity, and written documents and communication devices follow explicit reading, writing, and delivery processes. Each character thereby forms an independent understanding of the world, naturally giving rise to information asymmetry, misunderstanding, and gradual discovery.

Attention and memory retrieval operate dynamically around the current situation. During everyday activity, characters continuously assimilate routine environmental updates in the background; when salient external cues emerge or a current action concludes, a selective attention mechanism is triggered. Long-term memory comprises episodic experiences, semantic beliefs~\cite{tulving2002episodic}, and prospective intentions. Semantic beliefs record information provenance, enabling characters to evaluate credibility and revise prior assumptions when confronted with new evidence. Episodic experiences and prospective intentions are retrieved on demand according to temporal recency, prior consolidation, and relevance to current situational cues~\cite{conway2000construction}, bringing relevant past experiences into the active decision context.

Based on current perception, retrieved memories, and active goals~\cite{park2023generative}, the autonomous decision-making module selects the character's next course of action. High-level reflection and planning are modeled as cognitive processes that unfold over simulated time to update the character's internal intentions and beliefs. Depending on situational complexity, a character may think while continuing routine actions or pause current actions to focus on reasoning, adopting the new cognitive decision upon completion of the thinking period.

\subsection{Embodied and Multi-Agent Interaction}

Decisions formed internally by a character translate into concrete actions in three-dimensional space through a unified embodied interaction mechanism that combines open-ended action proposals~\cite{vezhnevets2023generative} with deterministic physical validation. Prior to decision-making, an affordance module supplies compact, context-relevant routine options for entities within the character's current focus of attention while preserving the ability to propose open-ended actions. Following action selection, the system validates physical preconditions against the live spatial state and compiles the chosen action into staged spatial and state updates. When an interaction encounters a physical obstruction during execution, the actual attempt and blocking cause return to the character as sensory feedback, driving adjustments to subsequent plans.

Interactions among multiple agents operate through the same mechanism. Verbal communication between characters and behavioral interactions over the shared environment proceed through explicit state updates and sensory delivery, allowing each participating character to retain independent decision-making authority within its own cognitive scope and to develop continuously evolving social relationships around shared spaces and objects.

\subsection{Persistent History and Narrative Discovery}

As the simulation progresses, all state changes, interaction events, character dialogues, and intention updates are recorded in chronological order within a persistent store. The storage system preserves both the objective world record and each character's personal observation history, supporting object state queries and event exports at any point in time from either a global objective viewpoint or a specific character's personal perspective. The episodic memories, semantic beliefs, and interpersonal relationships accumulated by a character in one simulation run can also be exported to initialize subsequent scenes, enabling long-horizon social evolution across scenes.

Open-ended social simulation generates abundant everyday activity while continually giving rise to pivotal dramatic turns~\cite{ryan2018curating}. Operating asynchronously in the background over the simulation history, the narrative discovery module filters out minor everyday activities and analyzes objective events, character dialogues, state transitions, and intention shifts to identify interconnected event arcs~\cite{kreminski2019felt}. The module condenses these developments into structured story checkpoints that capture key transitions including commitment formation, obstructed actions, interpersonal conflicts, and goal fulfillment, while retaining references to the underlying simulation events so that discovered stories remain faithful to actual history.

By specifying different target characters, themes, or viewpoints, the same simulation history yields multiple distinct scripts and storyboards that reflect how different individuals uniquely experienced and interpreted the same events, providing structured narrative inputs alongside the associated world state for downstream executable scene construction.
\section{Explicit State Control and Coding-Agent Orchestration}
\label{sec:explicit-state-control}

\begin{figure*}[t]
    \centering
    \includegraphics[width=\textwidth]{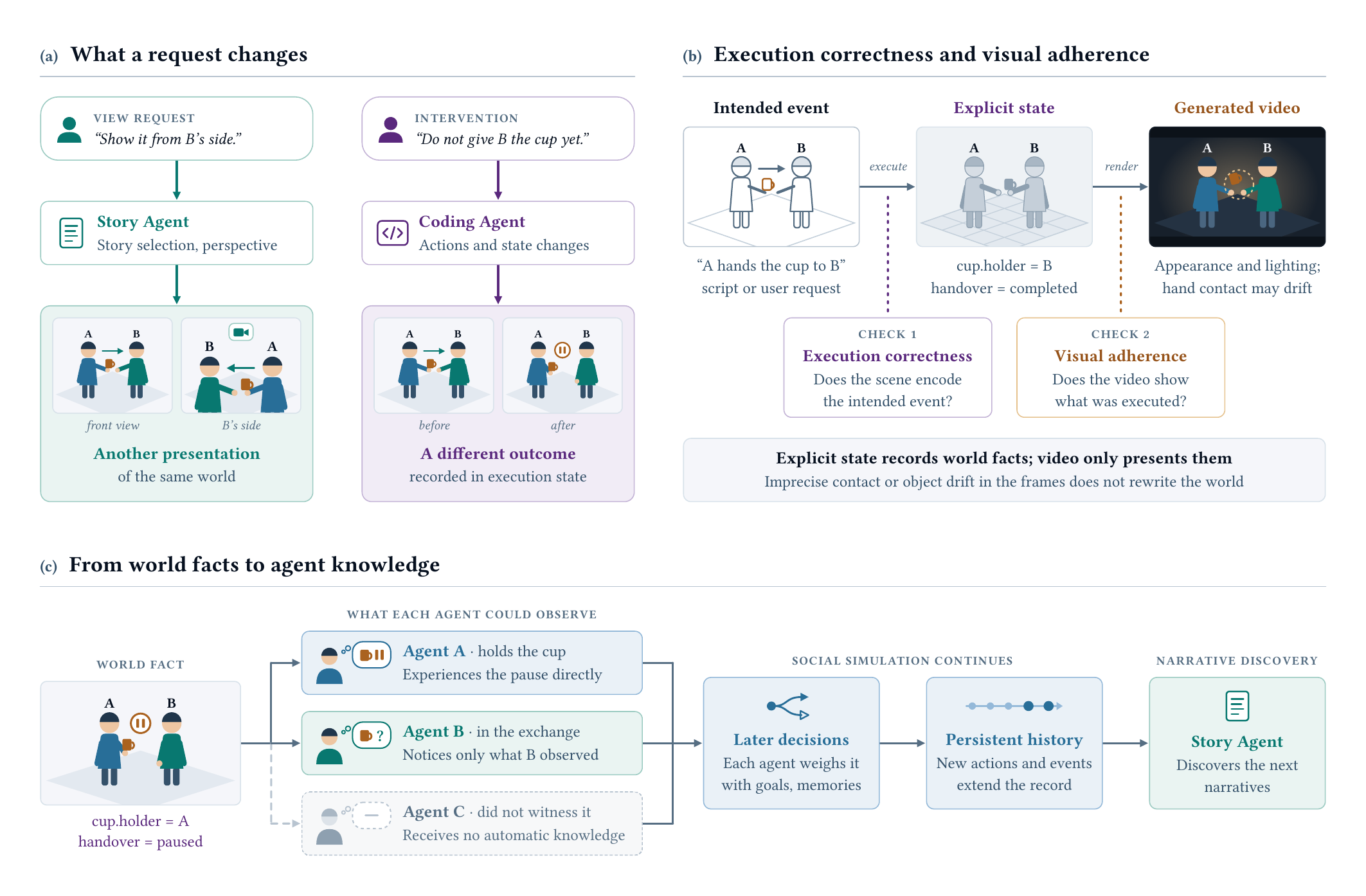}
    \caption{
    \textbf{Request semantics, execution fidelity, and
    observation-mediated social feedback.}
    (a) What a request changes.
    Viewpoint requests are handled at the narrative-selection and
    perspective level, providing another presentation of the same world.
    World-changing interventions are translated by the Coding Agent
    into actions and state changes, such as pausing a handover while
    A retains the cup.
    (b) Execution correctness and visual adherence.
    Two distinct checks compare the intended event with the explicit
    scene state, and the generated video with what was executed,
    respectively. Explicit state records world facts; rendering
    artifacts such as imprecise contact or object drift do not by
    themselves alter those facts.
    (c) From world facts to agent knowledge.
    The feedback design conditions subsequent social evolution on
    what each agent could observe: A experiences the pause directly,
    B receives only the information available from B's observations,
    and an absent C gains no automatic knowledge of the event.
    These observations inform later decisions together with goals
    and memories, extending the event history from which the
    Story Agent can discover subsequent narratives.
}
    \label{fig:explicit-state-control}
\end{figure*}

Social simulation describes why characters act, while cinematic rendering
determines how their actions are presented. Connecting these two processes
requires a world representation that can be maintained, executed, and modified.
A conversation, an object exchange, or a conflict has both social meaning and
concrete consequences: characters occupy particular locations, objects belong
to or are held by specific entities, and actions unfold in a particular order.
Making these properties explicit provides a foundation for executable scenes
and for user interventions with traceable consequences.

Related foundations include memory-based social agents~\cite{park2023generative}
and the separation of executable world evolution from visual realization in
Code World Model~\cite{chen2026code}. Zoooo-WM connects these concerns
through a social-to-execution interface and a feedback design for user intervention.

Zoooo-WM connects social simulation and video rendering through explicit scene
state and Coding-Agent orchestration. The Story Agent selects events from the
simulation history and organizes them into scripts and storyboards. The Coding
Agent translates these narratives, their associated world state, and user
requests into executable character actions, object interactions, and camera
trajectories. The runtime maintains scene state and produces the structural
conditions used by the video renderer. The interaction loop is designed to
return execution outcomes to social simulation through agents' observations
and experiences, allowing interventions to inform subsequent behavior and
story development.

\subsection{Social State and Execution State}
\label{subsec:social-execution-state}

Social state and execution state describe complementary aspects of the same
world. Social state includes agents' goals, memories, relationships, available
knowledge, and event histories. It provides the context in which an action
becomes meaningful. For example, character A may offer a drink to character B
as a gesture of goodwill, while B's response may depend on their relationship,
current goals, and previous interactions. These properties explain the event
but do not yet specify how it should unfold in a scene.

Execution state describes its concrete realization. It includes character
positions and orientations, active actions, object placement and attachment,
and camera configuration. Executing the drink handover requires specifying
where A and B are, who initially holds the cup, when the characters approach
each other, and how the cup's holding relation changes when the exchange
completes.

The two representations are connected through corresponding characters,
objects, and events. A narrative reference to character A must resolve to a
particular scene entity, and the cup mentioned in the event must correspond
to a trackable object. The same interaction can therefore be represented as
a socially meaningful exchange and as a concrete transition in an object's
holding relation.

Concordia similarly distinguishes agents' proposed actions from their effects:
agents describe intended actions, while a Game Master resolves their
implementation and consequences~\cite{vezhnevets2023generative}.
This separation also distinguishes intended actions from realized outcomes.
An intention to hand over the cup does not imply that the transfer has
occurred. Similarly, a user request to interrupt the exchange does not by
itself establish that the interruption succeeded. Subsequent state should
reflect the outcome of execution rather than only the outcome anticipated
by a script or request.

\subsection{From Narrative Intent to Executable Orchestration}
\label{subsec:executable-orchestration}

Program-based activity representations provide a precedent for translating
semantic tasks into executable actions: VirtualHome represents household
activities as sequences of atomic actions and interactions~\cite{puig2018virtualhome}.
Code as Policies further demonstrates language-model-generated programs that
compose control APIs from natural-language commands~\cite{liang2023code}.

The Story Agent selects events, viewpoints, scripts, and storyboards from
the evolving social history. The Coding Agent grounds these narrative
decisions in executable scenes. Its role is to determine which entities
participate, how their actions are organized, and how the resulting event
is presented.

This translation involves entity grounding, action coordination, and temporal
orchestration. Entity grounding associates narrative participants and props
with the current scene and its existing state. Scene construction can
therefore inherit relevant character locations, object relations, and event
context rather than redefining them independently for every segment.

Action coordination decomposes a semantic event into the behaviors needed
to realize it. For example, handing over a cup may involve approaching,
turning, extending an arm, receiving the object, and ending the interaction.
Some actions can overlap, while others depend on preceding actions or
conditions. Changes in object attachment must be coordinated with the
interaction rather than occurring independently of it.

Temporal orchestration places character actions, object transitions, and
camera motion on a shared timeline. A camera may follow a participant or
change viewpoint to emphasize an interaction, but these presentation choices
should not independently alter the event being depicted. Different shots
can present the same exchange, while the runtime determines whether the
exchange actually completes.

As illustrated in Figure~\ref{fig:explicit-state-control}a, the Coding Agent
produces scene code and action calls that are executed by a deterministic
runtime. Language-model reasoning handles semantic interpretation and
planning, while the runtime advances the scene according to explicit
execution logic. Determinism here concerns scene execution under specified
conditions; it does not imply that language-model planning or video generation
is itself deterministic.

The Coding Agent primarily translates narrative intent into execution rather
than independently rewriting character motivations or deciding the entire
subsequent story. When an intervention changes an event, the resulting
execution state provides new conditions from which social simulation can
continue.

\subsection{Connecting Execution State to Visual Rendering}
\label{subsec:execution-rendering}

Code World Model connects executable world state to video generation through
proxy videos encoding frame-wise spatiotemporal constraints~\cite{chen2026code}.
This provides a related formulation for separating world evolution from visual
realization.

The sequence of execution states is converted into frame-aligned structural
conditions for the video renderer. These conditions describe spatial layout,
motion, occlusion, and event timing. Appearance references and the reference
first frame specify visual identity and style, while text supplies additional
behavioral semantics and presentation requirements.

This division separates event facts from their visual realization.
Character entry, movement paths, and object exchanges can be specified at
the execution level. The renderer supplies appearance, lighting, expressions,
and motion detail conditioned on the available structure and references.
The extent of this generative freedom depends on the granularity of the
structural input and the renderer's ability to follow it.

Execution correctness and visual adherence are consequently distinct.
A scene may correctly encode a transfer of the cup from A to B while the
generated video exhibits inaccurate hand contact, object drift, or an
ambiguous exchange. Conversely, a visually plausible handover is not sufficient
evidence that the underlying execution state has been updated.

Maintaining explicit state and structural conditions makes these sources
of error distinguishable. The intended event can first be checked against
the executable scene, and its visual realization can then be compared with
the generated video. In this formulation, explicit state records world facts,
while the renderer presents those facts visually. Incidental changes in
generated imagery are not automatically treated as authoritative changes
to the world.

\subsection{Direct Control and Semantic Intervention}
\label{subsec:user-intervention}

Natural-language participation in simulated societies has been explored in
Generative Agents, where users can interact with agents in a persistent
sandbox environment~\cite{park2023generative}.
Users can participate through direct control or semantic intervention.
Direct control concerns character movement, camera control, and immediate
actions. Semantic intervention expresses a higher-level intention, such as
asking a character to leave, interrupting an exchange, or changing the
participants in an event. The Coding Agent translates world-changing requests
into executable actions or state changes.

Requests differ in their effects on the world. Selecting another story,
changing the viewpoint, or adjusting a camera primarily changes what is
shown and how it is presented. Pausing a handover, causing a participant to
leave, or introducing another character changes the conditions or outcomes
of an event. Narrative-selection requests are handled by the Story Agent,
while changes to the executable world must be reflected in scene execution.

An intervention does not assign the Coding Agent responsibility for every
downstream narrative consequence. For example, the request ``Do not give B
the cup yet'' can be realized by pausing the exchange and preserving A's
holding relation. Whether B becomes confused, asks a question, or changes
a later decision depends on B's observations, goals, and social context.
These responses belong to the continuation of social simulation.

Changing a story therefore requires more than generating a clip that depicts
a new instruction. The changed outcome must also become available to the
process that determines subsequent events. If an exchange is interrupted
in the scene but later decisions still assume that B received the cup,
the intervention has not produced a consistent, persistent consequence.

This interaction model can operate at event boundaries without requiring
continuous keyboard-level control. A user can intervene at a meaningful
moment, allow the system to orchestrate and execute the request, and then
observe the continuation. Its central property is that an action can change
the world and affect what follows.

\subsection{Checkpoints, State Persistence, and Branching}
\label{subsec:checkpoint-branching}

Checkpoints provide a shared starting point from which execution can continue
under different user choices. Branches originating from the same checkpoint
share the preceding entities, object states, and event context, allowing
their subsequent differences to be associated with the intervention.

Figure~\ref{fig:explicit-state-control}b illustrates this mechanism using a
cup handover. At the shared checkpoint, A holds the cup and the exchange is
pending. In the original branch, the event continues and the cup passes
to B. In the intervention branch, the exchange is paused and A retains
the cup. The important difference is a change in execution outcome, not
merely a difference between two images or descriptions.

State persistence also requires an intervention to have a meaningful scope.
Pausing an exchange should not arbitrarily change character identities,
reset the scene, or modify unrelated objects. Likewise, changing the camera
should not erase established relationships or previously completed events.
Preserving unaffected state makes the consequences of a local action
understandable and supports continuity across subsequent scenes.

A checkpoint intended to support continued social evolution must capture
more than the positions visible in a single frame. Relevant event progress
and associated social context must also be recoverable. Restoring these
conditions consistently is necessary for branches to differ because of
the intervention rather than because unrelated memories or object relations
were changed during restoration.

The branching example specifies the state-transition mechanism. The two
execution outcomes provide different conditions for later character
responses; the responses themselves are determined by the subsequent
social simulation.

\subsection{From Execution Outcomes to Social Consequences}
\label{subsec:execution-feedback}

The interaction loop is designed to return state changes and interaction
outcomes to social simulation. This feedback concerns both what happened
and what individual agents could observe.

World facts and agent knowledge are not identical. The cup remaining with A
is an execution outcome. Whether B notices the interruption or understands
A's intention depends on B's observations and prior information. Other
characters who did not witness the exchange should not automatically receive
complete knowledge of it.

The role of accumulated experience in subsequent behavior is supported by
Generative Agents, whose architecture integrates memory, reflection, and
planning~\cite{park2023generative}.
Execution outcomes must therefore enter the social process through
agent-specific observations and experiences. Different agents may interpret
the same event differently and combine it with their goals and memories
when making later decisions. Those decisions generate further actions
and events, extending the persistent history from which the Story Agent
can discover subsequent narratives.

This connection gives explicit control a role beyond local scene editing.
Users can affect immediate actions and, through their outcomes, change
the conditions under which later events develop. Explicit state supplies
traceable world facts, the Coding Agent connects semantic intent to scene
execution, and social simulation provides the mechanism through which
these changes can acquire persistent behavioral and narrative consequences.

\section{Cinematic Rendering and Streaming Execution}
Zoooo-WM renders the structured scene sequence into cinematic video using a visual first frame, text prompts, and per-frame structural guidance. The generated video is streamed incrementally in short chunks.

\subsection{Appearance Conditioning and First-Frame Construction}
We synthesize the visual first frame using an image generation model~\cite{openai2026images25} conditioned jointly on character, scene, and prop references and the first structure frame. The references specify the intended appearance, whereas the structure frame constrains spatial layout and composition. The resulting image serves as the appearance reference for subsequent structure-conditioned video generation.

\subsection{Structure-Conditioned Cinematic Rendering}
We support two structure-conditioned video backbones built on pretrained diffusion transformers ~\cite{peebles2023scalable,minimaxh3} in a compressed latent space: a bidirectional model and its causal streaming counterpart. Both backbones condition on the visual first frame, text prompts, and frame-aligned depth and semantic maps to jointly govern scene layout, motion, appearance, and semantic context.

\begin{figure}[t]
    \centering
    \includegraphics[width=1\linewidth]{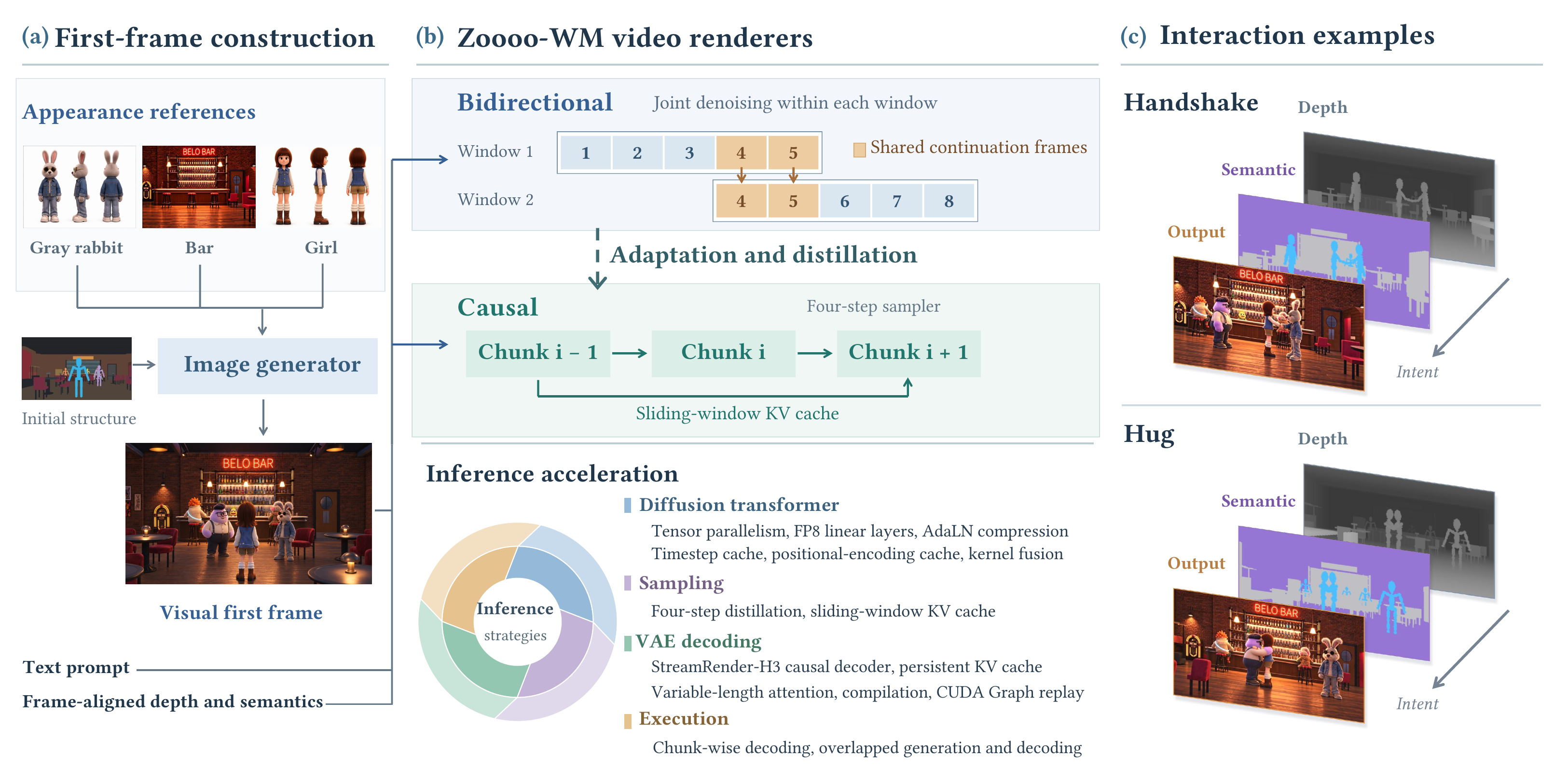}
   \caption{\textbf{Cinematic rendering and streaming execution in Zoooo-WM.}
(a) An image generation model constructs the visual first frame from
appearance references and the initial structure frame.
(b) Both renderers generate video using the visual first frame, text prompts,
and frame-aligned depth and semantic maps. The bidirectional model jointly denoises
frames within each window and uses overlapping frames for continuation,
whereas the causal streaming model generates chunks using four-step
sampling and a sliding-window KV cache. Streaming inference optimizations
span DiT computation, sampling, VAE decoding, and execution.
(c) Depth and semantic maps guide cinematic rendering in the handshake and hug examples.}
    \label{fig:structure_conditioned_rendering}
\end{figure}
The bidirectional model  \cite{chen2026code} jointly denoises
video latents within each temporal window, using bidirectional attention
and the complete structural conditions for that window. Longer sequences
are generated through overlapping windows. Each subsequent window conditions on the tail frames of the preceding window alongside the persistent appearance reference.

For low-latency interactive scenarios, the causal streaming model \cite{liao2026streamingcwm}, which is adapted and distilled from the bidirectional architecture, generates video latents autoregressively chunk by chunk via a four-step sampler. A sliding-window KV cache propagates recent context across adjacent chunks to maintain temporal coherence without future lookahead.

\subsection{Streaming Execution and Inference Acceleration}
The renderer processes structural conditions chunk by chunk and converts the generated video latents into RGB frames using a streaming VAE decoder \cite{streamrenderh3_2026}. The decoder reuses cached history to decode each completed chunk without waiting for future video latents. By pipelining the generation of chunk $t+1$ with the VAE decoding of chunk $t$, the system achieves low-latency incremental output.

We accelerate inference through tensor parallelism~\cite{shoeybi2019megatron}, FP8 linear layers, low-rank AdaLN compression, timestep and positional-encoding caches, and kernel fusion\cite{li2026solvideoinferenceengine}. VAE decoding uses StreamRender-H3's 24-layer distilled causal decoder~\cite{streamrenderh3_2026}, with persistent KV caching, variable-length attention, compilation, and CUDA Graph replay. This configuration generates $1344 \times 768$ video at 14.30 frames per second on eight H200 GPUs, emitting a 17-frame chunk every 1.189 seconds on average in steady state.
\section{Capabilities and Applications}

\subsection{A Comparative Perspective on Interactive Worlds}
\label{sec:comparative_interactive_worlds}

Table~\ref{tab:interactive_worlds_comparison} situates Zoooo-WM
alongside four related approaches to interactive worlds.
Representative examples include authored branching experiences such as
\textit{Bandersnatch} and Telltale's \textit{The Wolf Among Us}%
~\cite{nee2021wild,ortet2018avaliaccao};
interactive video world models such as the Genie,
LingBot-World, WorldPlay, and SolarWM%
~\cite{bruce2024genie,hadsell2026genie4,team2026advancing,sun2025worldplay,huang2026solarwm};
agent-based social simulation in Generative Agents%
~\cite{park2023generative};
and executable-state approaches such as Code World Model and
Programmable World Model%
~\cite{chen2026code,huang2026programmable}.

We compare these approaches along four complementary dimensions.
\textit{Memory and persistence} describes the information retained
over time, including choice history, visual context, world state,
and agent memories.
\textit{Agency and participation} concerns how characters act
and how users influence their behavior or the environment.
\textit{Narrative variation} describes how alternative experiences
arise, through authored branches, generated continuations, or
different stories discovered within a shared history.
\textit{State and rendering} concerns the mechanisms that
determine world changes and translate them into visual observations.

\begin{table}[htbp]
    \centering
    \caption{Qualitative comparison of representative interactive-world
    designs.}
    \label{tab:interactive_worlds_comparison}

    \small
    \setlength{\tabcolsep}{4pt}
    \renewcommand{\arraystretch}{1.15}

    \begin{tabularx}{\linewidth}{
        *{5}{>{\raggedright\arraybackslash}X}
    }
        \toprule
        {\bfseries Approach}
        & {\bfseries Memory \& persistence}
        & {\bfseries Agency \& participation}
        & {\bfseries Narrative variation}
        & {\bfseries State \& rendering}
        \\
        \midrule

        Authored Branching Narratives~\cite{nee2021wild,ortet2018avaliaccao}
        & Authored variables and choice history
        & Scripted behavior; choices and gameplay
        & Prepared branches; new outcomes require authoring
        & Authored logic; recorded or engine-rendered visuals
        \\
        \midrule[0.25pt]

        Interactive Video World Models~\cite{bruce2024genie,hadsell2026genie4,team2026advancing,sun2025worldplay,huang2026solarwm}
        & Video context; model-dependent memory
        & Action- or camera-conditioned interaction
        & Generated continuations under user control
        & Learned dynamics; generated visual continuations
        \\
        \midrule[0.25pt]

        Agent-based Social Simulation~\cite{park2023generative}
        & World state; agent memories and plans
        & Autonomous agents; communication and user influence
        & Emergent events and social trajectories
        & Agent-driven environment; authored visual interface
        \\
        \midrule[0.25pt]

        Executable-state Generative Worlds~\cite{chen2026code,huang2026programmable}
        & Persistent entities, attributes, and rules
        & Code-orchestrated interactions and user control
        & Alternative trajectories via state or code changes
        & Executable state; generative visual rendering
        \\
        \midrule

        {\bfseries Zoooo-WM (Ours)}
        & World records; agent memories and relationships
        & Autonomous agents; direct and semantic intervention
        & Stories selected from shared history; checkpoint branches
        & Executable state; generative cinematic rendering
        \\
        \bottomrule
    \end{tabularx}
\end{table}

The comparison highlights different ways of organizing continuity
and participation. 
Authored branching narratives provide deliberate authorial control,
but their narrative possibilities remain bounded by prepared content
and choice logic. Character responses and lasting consequences depend
on the situations anticipated by their authors; extending the experience
to substantially new outcomes requires additional writing, production,
and validation. This creates an authoring burden when moving from
a finite set of narrative paths toward continuously evolving
character relationships and stories.

Interactive video world models expand beyond prepared visual content
through action-conditioned generation, with recent advances improving
memory and long-horizon consistency%
~\cite{sun2025worldplay,huang2026solarwm}.
However, maintaining a visually consistent environment does not by
itself establish persistent records of what each character knows,
remembers, or intends. Where generated observations become context for
subsequent prediction, visual errors can also influence later
continuations. Precise social consequences therefore remain difficult
to inspect and enforce through visual generation alone.

Agent-based social simulation directly addresses autonomous behavior
and evolving relationships through agent memories, reflection, and
planning~\cite{park2023generative}.
Its remaining challenge for cinematic experiences is that emergent
activity does not automatically form a compelling, selectively
presented story. Events may be dispersed across characters and time,
requiring narrative discovery to identify meaningful arcs and
viewpoints~\cite{leong2022automated}.
Moreover, the authored visual interface of systems such as Smallville~\cite{park2023generative}
does not itself provide cinematic realization of these evolving events.

Executable-state generative worlds separate programmable world
evolution from visual generation, supporting persistent entities,
rules, and controllable interactions%
~\cite{chen2026code,huang2026programmable}.
Their demonstrated emphasis is on constructing and controlling
interactive environments; organizing an ongoing society into
selected, character-centered stories remains an additional task.
Explicit state can represent goals and social consequences, but
does not by itself provide the orchestration needed to connect
autonomous social evolution, narrative discovery, and cinematic
participation.

Zoooo-WM brings these requirements together through long-horizon
social evolution, multi-agent interaction, and explicit state control.
A persistent history of events, individual experiences, and
relationships provides reusable material from which different
stories and viewpoints can be discovered. Direct and semantic
interventions further support alternative continuations from shared
checkpoints, allowing users to influence events beyond selecting
among prepared narrative branches.
Maintaining world state separately from rendered video
also makes execution correctness and visual fidelity distinct
questions: incidental visual changes are not automatically committed
as world facts. Execution feedback is designed to carry outcomes
into subsequent agent experiences and decisions.
Combined with pretrained cinematic rendering, this organization
connects autonomous story development with user participation,
supporting story exploration and visualization for filmmaking,
as well as playable, character-driven experiences for games and
interactive cinema.

\subsection{Explicit State Control and Alternative Continuations}
\label{sec:explicit_control_continuations}

\begin{figure}[p]
    \centering
    \includegraphics[
        width=\linewidth,
        height=0.88\textheight,
        keepaspectratio
    ]{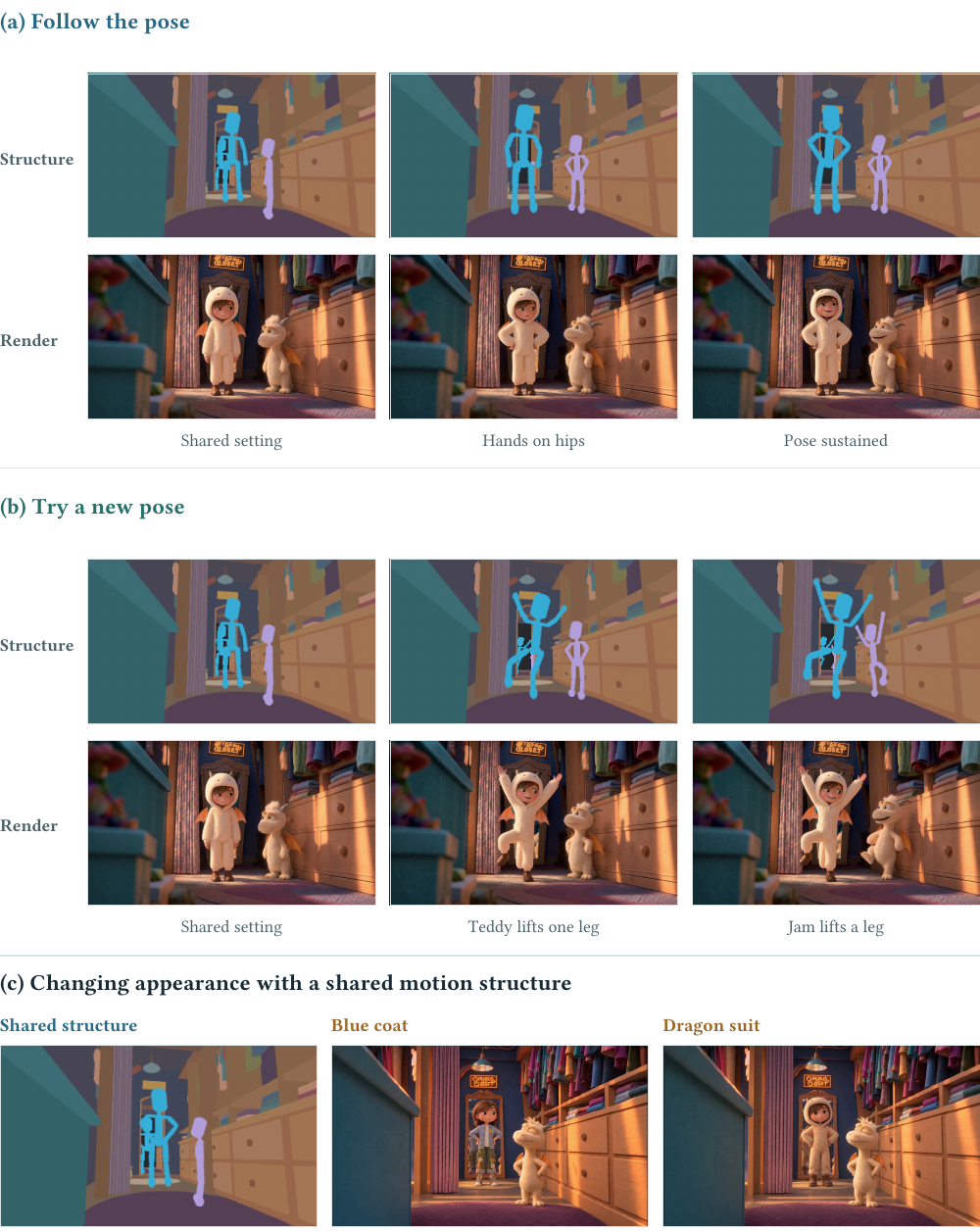}
    \caption{\textbf{Explicit control of actions and appearance.}
    (a,b) Corresponding frames from the structure and rendered videos show
two continuations: a hands-on-hips pose, and Teddy lifting one leg
followed by Jam lifting a leg.
(c) A shared motion structure guides two appearance variants:
a blue coat and a dragon suit.}
    \label{fig:control_appearance}
\end{figure}

We illustrate Zoooo-WM's capabilities through three scenes that connect
controllable behavior, multi-agent interaction, and cinematic presentation.
The fitting-room example examines how an action changes an exchange between
Teddy (a young girl) and Jam (a small dragon)
(Figure~\ref{fig:control_appearance}(a,b)). In one
continuation, Teddy wears a dragon suit and takes a hands-on-hips pose beside
Jam. In another, she raises both arms and balances on one leg; Jam
subsequently lifts a leg as well. The contrast extends beyond a change in
one character's pose to the visible response that follows it. Each branch shows three pairs of corresponding frames from the structure
and rendered videos, making the prescribed actions and their cinematic
realization directly comparable.

A separate comparison retains the same motion structure while rendering
Teddy in either a blue coat or a dragon suit
(Figure~\ref{fig:control_appearance}(c)). The comparison varies the outfit
around a shared motion specification and a similar spatial arrangement.
Together, these examples distinguish control over what characters do from
control over how they appear. Executable structure provides an inspectable
description of the intended behavior, while appearance conditioning allows
the visual design to vary around it.

This separation makes \textit{explicit state control} useful for creative
iteration. Filmmakers can explore performance and costume alternatives
within a shared scene, using the structure as a reference for the intended
action. In games and interactive cinema, changing an action can shape the
ensuing character exchange, extending participation beyond navigation.
Zoooo-WM thus connects editable behavior to cinematic variations that
creators can compare and refine.

\subsection{Multi-Agent Interaction and Narrative Outcomes}
\label{sec:multi_agent_narrative_outcomes}

\begin{figure}[!htbp]
    \centering
    \includegraphics[width=\linewidth]{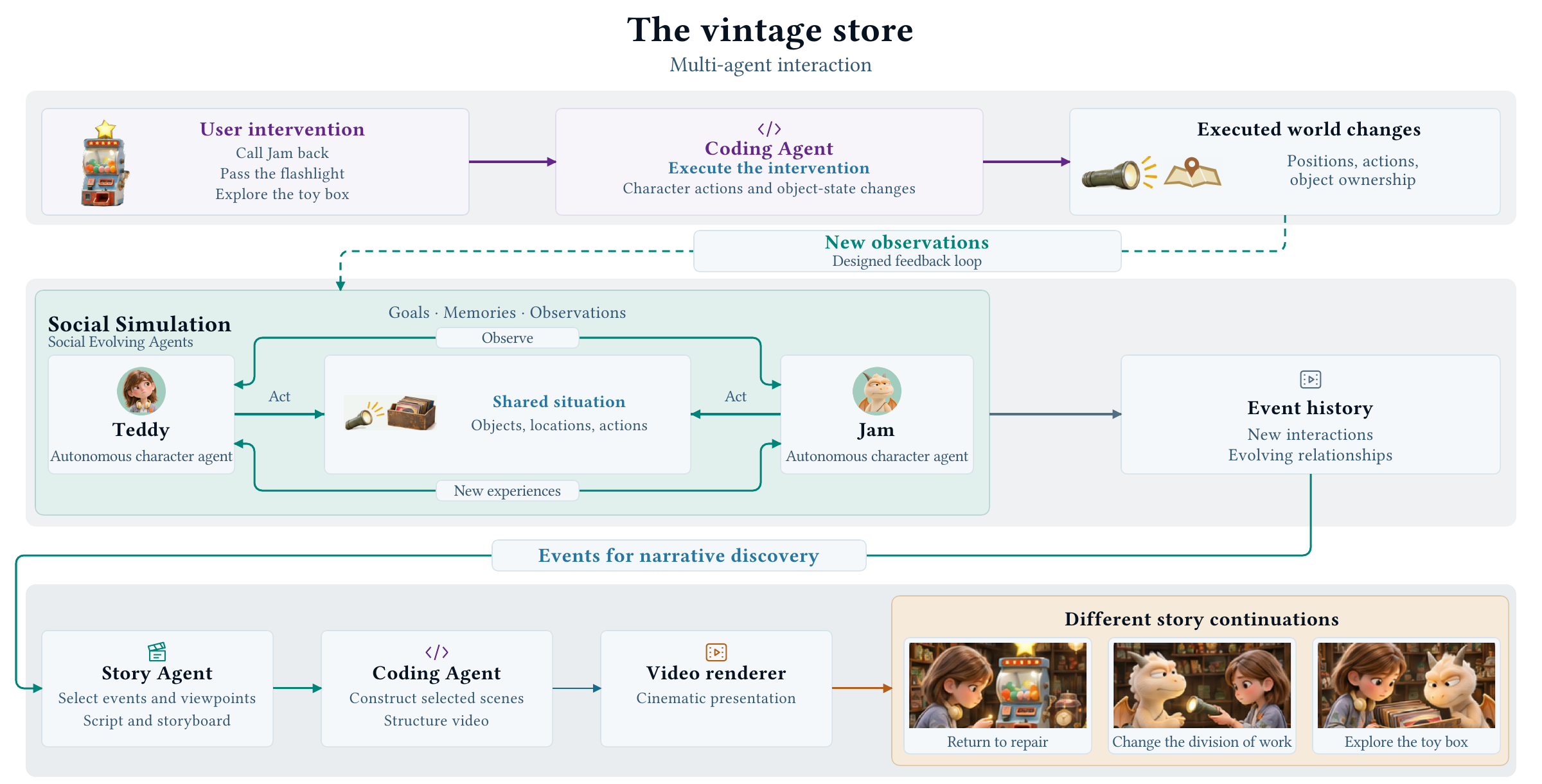}
    \caption{\textbf{Multi-agent interaction in the vintage store case.}
    Interventions change a shared situation and its possible continuations.
    The designed feedback pathway connects executed changes, agent
    observations, and evolving event history to subsequent narrative discovery.}
    \label{fig:vintage_interaction}
\end{figure}

The vintage-store example places multi-agent interaction around a shared
task. After a power outage, Teddy attends to a capsule machine, Jam holds
a flashlight, and a nearby toy box offers a competing activity.
Figure~\ref{fig:vintage_interaction} relates this setting to Zoooo-WM's
interaction and narrative-discovery mechanisms.

\begin{figure}[!htbp]
    \centering
    \includegraphics[width=\linewidth]{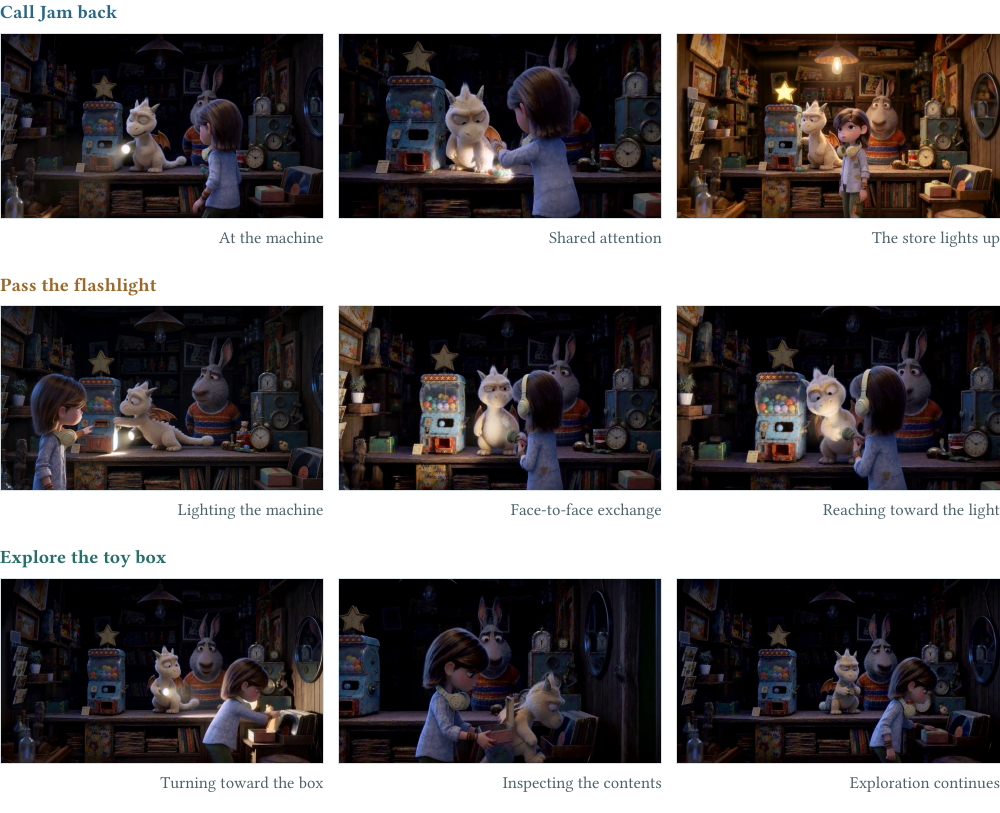}
    \caption{\textbf{Different interactions, different story continuations.}
    Calling Jam back returns attention to the machine; the flashlight
    exchange brings the characters face to face; exploring the toy box
    redirects their activity.}
    \label{fig:vintage_continuations}
\end{figure}

Different interventions produce distinct continuations
(Figure~\ref{fig:vintage_continuations}). Calling Jam back brings attention
to the machine again, and the sequence ends with the store illuminated.
The flashlight-exchange continuation brings the characters face to face,
with their gestures centered on the light. Exploring the toy box instead
redirects their activity toward its contents while the store remains dark.
These alternatives change the focus and organization of the shared
activity, producing recognizable differences in how the scene develops.

The distinguishing role of \textit{multi-agent interaction} is that actions
acquire meaning through their relation to other characters and the ongoing
situation. Returning to a task, engaging another character, and exploring
an object are different developments of the same social setting. Explicit
state control makes such changes actionable. The interaction between
characters gives them narrative significance beyond isolated movements.

These scene-level alternatives also clarify the role of
\textit{long-horizon social evolution} in Zoooo-WM. In the system design,
executed changes feed into agents' observations and experiences, and the
Story Agent selects connected events and viewpoints from an evolving
history. A cinematic episode is therefore a selected view of an ongoing
society, whose activity can continue beyond that episode. The case diagram
shows how local interventions are intended to connect to this broader
process.

For filmmaking, the case supports exploring alternative story developments
within a shared premise: a scene may return to cooperation or move toward
distraction and discovery. For games and interactive cinema, it illustrates
participation through changes to a common task and the characters'
involvement in it. The resulting variation concerns how a situation
unfolds, as well as how it looks.

\subsection{Context-Sensitive Social Interaction}
\label{sec:context_sensitive_social_interaction}

\begin{figure}[!htbp]
    \centering
    \includegraphics[width=\linewidth]{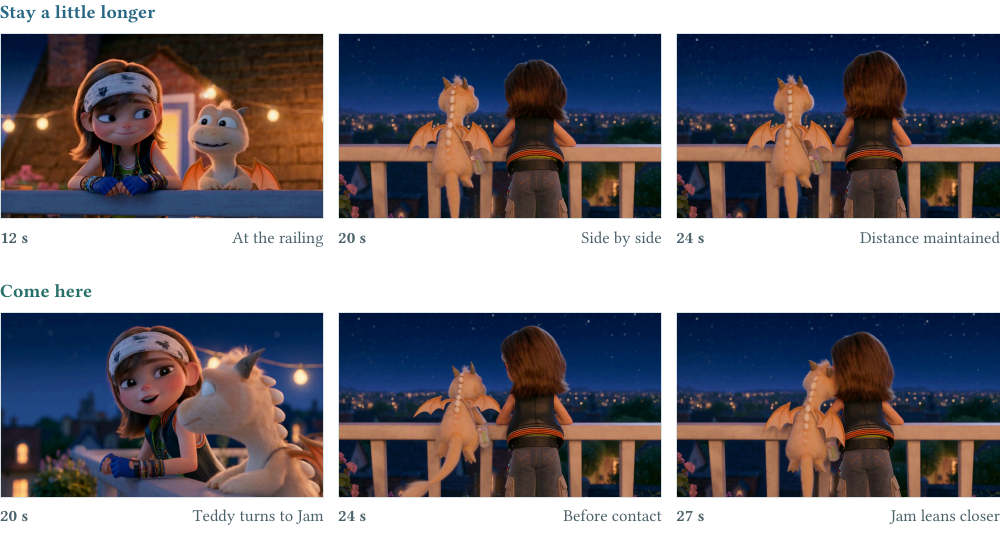}
    \caption{\textbf{Social responses through movement and proximity.}
    The \textit{Stay a little longer} continuation preserves space between the characters.
    Following an invitation in the \textit{Come here} continuation, Teddy turns toward Jam and Jam moves closer,
    ending in shoulder contact. Each row shows successive moments from its
    own video.}
    \label{fig:rooftop_interaction}
\end{figure}

The rooftop example examines how an invitation changes a shared moment
(Figure~\ref{fig:rooftop_interaction}). In the \textit{Stay a little longer}
continuation, Teddy and Jam remain at the railing with space between them.
In the \textit{Come here} continuation, Teddy turns toward Jam; Jam
subsequently moves closer and leans against Teddy's shoulder. The contrast
is expressed through orientation, the sequence of actions, and physical
proximity, rather than dialogue alone.

This example connects Zoooo-WM's \textit{playability and cinematic
presentation}. Participation can influence an encounter whose meaning is
carried by the characters' behavior. Structure conditioning specifies
movement and spatial relationships, while rendering gives the exchange
its visual appearance and atmosphere. The result is a socially meaningful
variation within a cinematic scene.

For filmmakers, these alternatives support exploring staging and emotional
beats through visible performance. For character-driven games and
interactive cinema, they suggest experiences in which user choices are
expressed through attention, companionship, and interpersonal distance.
Across the three cases, Zoooo-WM connects controllable actions with
character interaction and cinematic storytelling, providing a basis for
worlds that can be both watched and participated in.

\paragraph{Limitations.}
The current renderer does not always strictly adhere to the prescribed
structure, particularly during complex contact and object interactions.
Learning to balance structural adherence with the flexibility of diffusion
models to generate natural motion, behavior, and interaction remains an
important direction for future work. Maintaining visual fidelity and
preserving socially coherent consequences over extended interaction also
remain open challenges. Longer closed-loop evaluations are needed to
assess persistence, responsiveness, and the effects of intervention.
Expressive behavior in a rendered scene should also be distinguished from
demonstrated social or emotional understanding.
\section{Impact}
\label{sec:impact}

Zoooo-WM takes an initial step toward social world models in which
characters' histories, relationships, and reasons for acting
shape both how a world evolves and which experiences are
worth presenting.
By connecting this social context with executable interaction,
our work offers a basis for exploring embodied behavior that
accounts for what others know, what they seek, and how previous
encounters shape their expectations.
This motivates embodied AI that brings social and
emotional understanding alongside physical competence,
informing when to assist, how to cooperate, and when to
refrain from an otherwise feasible action.
For filmmaking and games, Zoooo-WM connects an evolving society
with the discovery and cinematic realization of meaningful
stories.
The same social history can support different perspectives,
while user participation opens possibilities for alternative
continuations.
This brings character and relationship development into
story creation and interactive play, supporting cinematic
experiences that audiences can both follow and influence.
\section{Conclusion}
\label{sec:conclusion}

Zoooo-WM explores social history as a shared foundation for
cinematic storytelling and play, connecting characters'
experiences with stories that users can follow and influence.
This work offers an initial direction toward social world
models that sustain evolving characters, relationships,
and meaningful consequences over time.
Establishing the coherence and reliability of these worlds
across extended interactions remains an important next step.

\paragraph{Future Directions.}
Moving toward richer social world models requires coupling physical
world state with models of beliefs, goals, emotions, and relationships.
One direction is to learn character-state updates from long-horizon
interaction trajectories, grounding these updates in each agent's
observations, persistent memory, and executed outcomes. This would
support modeling how characters interpret the same event differently
and how experiences influence subsequent decisions and relationships.
For visual realization, training on aligned structure and video
trajectories could improve controllability through objectives for
motion, contact, and object-state consistency, while preserving the
generative flexibility needed for natural behavior and interaction.
For narrative discovery, the Story Agent could learn event-selection
and viewpoint policies from human feedback, considering causal
connections, character development, and audience interest.
These directions call for closed-loop evaluation using replayable
checkpoints and controlled interventions, combining state-based
consistency checks with human assessment of character behavior,
emotional coherence, and narrative quality.
\section{Authors}
\label{sec:authors}

Zeyu Zhang, Xianghao Kong, Songlin Yang

Longji Huang, Mingchen Li, Xuyi Yang, Hohin Kwan, Shiqin Hou

Yudi Xie, Siyu Huang, Xuhuan Shen

Eddy Chen, Lvmin Zhang, Anyi Rao
\clearpage
\bibliographystyle{plainnat}
\bibliography{zoooo}

\end{document}